\documentclass{article}

\usepackage{amsmath, amssymb, graphicx}

\usepackage{hyperref}
\hypersetup{
    colorlinks=true,
    citecolor=brightblue,  
    linkcolor=black,       
    urlcolor=blue          
}

\usepackage{url}
\usepackage{wrapfig}
\usepackage[utf8]{inputenc} 
\usepackage[T1]{fontenc}    
\usepackage{tikz}
\usetikzlibrary{positioning}
\usepackage{booktabs}       
\usepackage{nicefrac}       
\usepackage{microtype}      
\usepackage{tabularx}
\usepackage{mathrsfs}
\usepackage{multirow}
\usepackage{array}
\usepackage{algorithm}
\usepackage{algorithmic}
\usepackage{subfigure}

\usepackage{hyperref}

\usepackage[accepted]{icml2025}

\usepackage{amsmath}
\usepackage{amssymb}
\usepackage{mathtools}
\usepackage{amsthm}
\usepackage{caption}
\usepackage{float}
\usepackage{enumitem}

\usepackage[capitalize,noabbrev]{cleveref}

\theoremstyle{plain}

\theoremstyle{definition}

\usepackage[textsize=tiny]{todonotes}

\usepackage{pifont} 

\icmltitlerunning{Mutual Information Preserving Neural Network Pruning}

\begin{document}

\twocolumn[
\icmltitle{Mutual Information Preserving Neural Network Pruning}



\icmlsetsymbol{equal}{*}

\begin{icmlauthorlist}
\icmlauthor{Charles Westphal}{yyy}
\icmlauthor{Stephen Hailes}{yyy}
\icmlauthor{Mirco Musolesi}{yyy,comp}

\end{icmlauthorlist}

\icmlaffiliation{yyy}{Department of Computer Science, University College London, London, UK}
\icmlaffiliation{comp}{Department of Computer Science, University of Bologna, Bologna, Italy}

\icmlcorrespondingauthor{Charles Westphal}{charles.westphal.21@ucl.ac.uk}

\icmlkeywords{Machine Learning, ICML}

\vskip 0.3in
]



\printAffiliationsAndNotice{} 
\begin{abstract}
Pruning has emerged as the primary approach used to limit the resource requirements of large neural networks (NNs). Since the proposal of the lottery ticket hypothesis, researchers have focused either on pruning at initialization or after training. However, recent theoretical findings have shown that the sample efficiency of robust pruned models is proportional to the mutual information (MI) between the pruning masks and the model's training datasets, \textit{whether at initialization or after training}. In this paper, starting from these results, we introduce Mutual Information Preserving Pruning (MIPP),  a structured activation-based pruning technique applicable before or after training.  The core principle of MIPP is to select nodes in a way that conserves MI shared between the activations of adjacent layers, and consequently between the data and masks. Approaching the pruning problem in this manner means we can prove that there exists a function that can map the pruned upstream layer's activations to the downstream layer's, implying re-trainability. We demonstrate that MIPP consistently outperforms state-of-the-art methods, regardless of whether pruning is performed before or after training. 
\end{abstract}
\section{Introduction}
\begin{figure}[t]
    \vskip 0.2in
    \centering
    \includegraphics[width=\columnwidth]{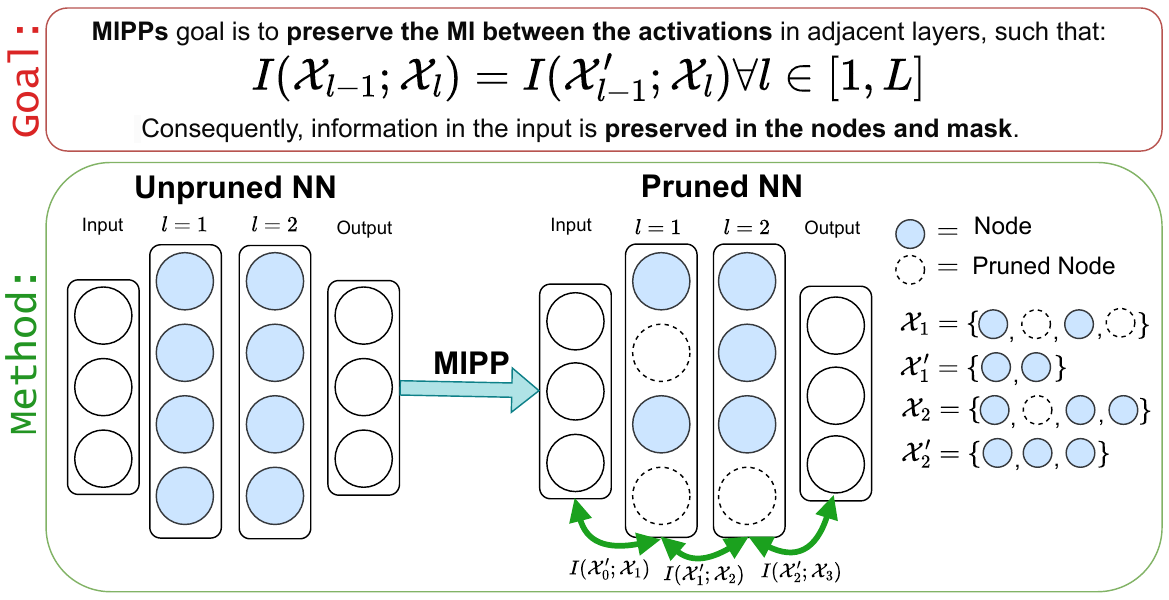}
    \caption{We introduce MIPP via an illustration. MIPP acts to preserve the mutual information (MI) between the activations in adjacent layers. In turn, this leads to a pruned network representation whose nodes and mask replicate the information contained in the data.}
    \label{fig:schem}
    \vskip -0.2in
\end{figure}
\begin{figure*}[t]
    \vskip 0.2in
    \centering
    \includegraphics[width=\textwidth]{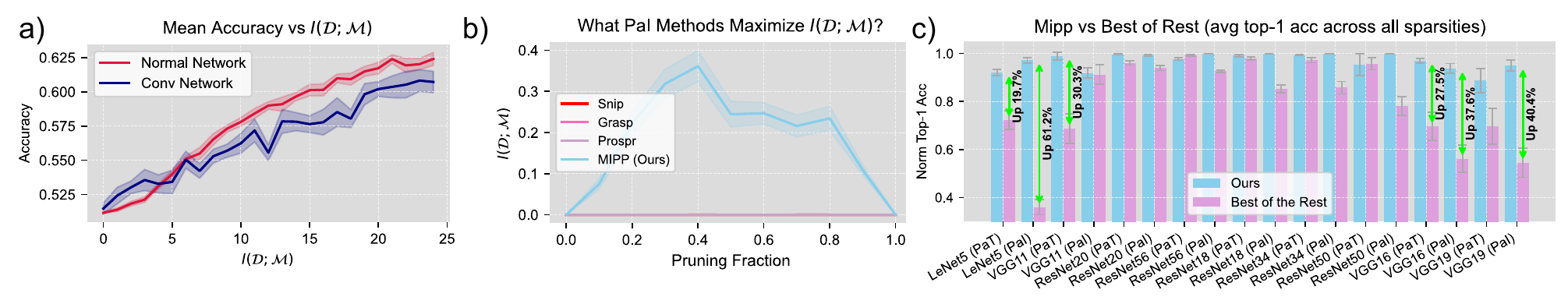}
    \caption{\textit{a)} Graphical representation of how MI between the mask and the data affects the accuracy of a small convolution-based and standard NN: we observe that by maximizing MI, the classification accuracy increases. The experiments are based on synthetic data; for full details refer to Appendix \ref{app:linei}. \textit{b)} A study examining how pruning masks, created using various PaI methods and applied to a small synthetic network, affect the values of \( I(\mathcal{D};\mathcal{M}) \). For full details about these experiments, please refer to Appendix \ref{app:whatpai}. \textit{c)} Comparison of MIPP's average accuracy across different sparsity ratios to the best-performing baseline for each model-dataset combination. MIPP outperforms the best of the rest significantly, as at high sparsities, they are all much more prone to layer collapse. PaT baselines: OTO, IMP, SOSP-H, ThiNet. PaI baselines: IterSNIP, IterGrasP, ProsPr, SynFlow.}
    \label{fig:obj}
    \vskip -0.2in
\end{figure*}
It is well-established that to limit a model’s resource requirements while maintaining its performance, it is preferable to \textit{prune} and re-train a large model of high accuracy rather than train a smaller model from scratch \citep{lecun1989optimal,lecun1998gradient,li2017pruning,han2015learning}. The lottery ticket hypothesis demonstrated that this was due to the existence of performant dense sub-networks embedded in overparameterized models at initialization \cite{frankle2018lottery}. This discovery motivated a new body of research on \textit{pruning at initialization (PaI)}, such as SNIP \citep{lee2019snip}, GraSP \citep{wang2020picking}, SynFlow \citep{tanaka2020pruning}, and ProsPr \citep{alizadeh2022prospect} to name a few. Sub-networks identified using these methods perform worse than those obtained through \textit{pruning after training (PaT)}, even when using straightforward approaches like iterative magnitude pruning (IMP) \cite{frankle2020pruning}. \citet{kumar2024no}'s PAC-learnability result provided an information-theoretic justification for this, demonstrating that the sample efficiency of a pruned learning algorithm is proportional to the effective parameter count, which can be calculated by summing the number of unmasked parameters and the mutual information (MI) shared between the pruning mask and training data. Based on this relationship, they posit that to maximize sample efficiency, it is essential for training to occur.

To provide an intuition for this PAC-learnability result by \citet{kumar2024no}, Figure \ref{fig:obj}.a illustrates the improved accuracy that results from maximizing MI between the pruning mask and the training data in both a standard neural network (NN) and a convolutional variant. For these experiments, the data were synthetic and masks derived before training, so MI values were obtained analytically. Given the result presented in \cite{kumar2024no}, supported in part by Figure \ref{fig:obj}.a, we conclude that maximizing MI shared between the pruning mask and training data is a sensible objective when pruning.

As originally argued by \citet{kumar2024no}, optimizing this objective is expected to restrict us to a PaT approach; without training, we have no reason to expect that the weights or pruning masks will exhibit any correlation with the data. While this certainly holds for data-independent pruning schemes, such as magnitude PaT or other PaI solutions like those presented by \citet{tanaka2020pruning,patil2021phew,pham2024towards}, it may not be universally true. For example, consider taking an activation-based approach. The activations at each layer of a NN are a function of the activations preceding them, or of the input data. If the NN is sufficiently expressive, these activations should contain all the information in the data, whether training has occurred or not. Therefore, if we can define a mask that preserves all the information in the activations, it should transfer to the data and maximize our objective, even at initialization.

Consequently, we introduce Mutual Information Preserving Pruning (MIPP), a structured activation-based pruning technique applicable before or after training. MIPP ensures that MI shared between activations in adjacent layers is preserved during pruning (please refer to Figure \ref{fig:schem}). Rather than ranking nodes and selecting the top-$k$, MIPP uses the transfer entropy redundancy criterion (TERC) \citep{westphal24} to dynamically prune nodes whose activations do not transfer entropy to the downstream layer. We will show that pruning in this manner ensures the existence of a function that can re-construct the downstream layer from the pruned upstream layer. Moreover, we will demonstrate that MIPP establishes pruning masks whose MI with the training data has a maximal upper bound. This is because MIPP dynamically evaluates and removes redundant nodes in a manner dependent on those currently maintained in the network representation, a feature that is unachievable using static ranking-based pruning methods. To illustrate this visually, in Figure \ref{fig:obj}.b we show that only MIPP can derive useful pruning masks for a synthetic NN characterized by nodes sharing redundant information. Finally, we demonstrate MIPP's utility beyond theoretical justification by presenting improved pruning results in both post- and pre-training domains, as shown in Figure \ref{fig:obj}.c. To summarize, the contributions of this work are as follows: 
\begin{itemize}[noitemsep, topsep=0pt]
\item We develop MIPP, a structured activation-based pruning method that preserves MI between the activations of adjacent layers in a deep NN. 
\item We prove that perfect MI preservation ensures the existence of a function, discoverable by gradient descent, which can approximate the activations of the downstream layer from the activations of the preceding pruned layer. Consequently, MIPP implies re-trainability.
\item We prove that pruning using MIPP leads to a maximum upper bound on MI between the data distribution and the mask distribution (as defined in \citet{kumar2024no}).
\item Through comprehensive experimental evaluation, we demonstrate that MIPP can effectively prune networks, whether they are trained or not. \end{itemize}

\section{Related Work}\label{sec:relwork}
\textbf{Pruning after training.} Traditional structured pruning methods employ metrics such as weight magnitude \citep{han2015learning,li2017pruning,wang2023trainability}, weight gradient \citep{lecun1998gradient,molchanov2017pruning}, Hessian matrices \citep{hassibi1992second,peng2019collaborative,wang2019eigendamage,nonnemacher2022sosp}, and combinations thereof, to rank and then remove nodes up to a defined pruning ratio (PR). Although these methods were originally designed to be applied at the level of individual weights, they can be adapted for structured cases through non-lossy functions, such as L1-normalization \cite{wang2023trainability}. This can be carried out in either a global or local manner, the former involves ranking all the nodes in a network \citep{liu2017learning,wang2019eigendamage}, while the latter is only applied to individual layers \citep{zhao2019variational,sung2024ecoflap}. Global methods have been effective in determining layer-wise pruning ratios \citep{blalock2020state}.  However, at high PRs, they experience layer-collapse, an undesirable final result in which an entire layer is pruned and an untrainable network is produced \cite{tanaka2020pruning}. Traditional methods, including magnitude, gradient, and Hessian-based approaches, continue to represent the state-of-the-art (SOTA) due to recent methodological refinements. Modern variations of such techniques are \textit{iterative}, meaning that the model is trained, some fraction of the weights - lower than the final PR - are removed according to the methods described, and then the model is retrained and the process is repeated until the PR is reached \cite{frankle2018lottery,you2020drawing}. These methods are known to lead to highly performant models, while also being resistant to layer collapse. However, they are computationally expensive because they require multiple retraining sessions. In response, methods such as SOSP-H have been proposed \cite{nonnemacher2022sosp}. SOSP-H ranks and removes nodes in a traditional way, except for the fact that the metric employed is the Hessian \cite{hassibi1992second}. The Hessian is recognized as the most computationally expensive yet best-performing metric \cite{molchanov2019importance}. By employing a second-order approximation, its benefits can be leveraged in a computationally efficient manner. 
While MIPP acts globally, aligning with the methods discussed thus far, it is also activation-based, diverging from these competing techniques. ThiNet \cite{luo2017thinet} most closely resembles MIPP in terms of methodology, although it is known that it is unable to establish layer-wise PRs, which ensures an inability to conduct any neural architecture search \cite{patil2021phew}.

\textbf{Pruning at initialization.} In contrast to pruning after training, pruning at initialization aims to identify and remove redundant parameters before the training process begins, thereby reducing computational overhead from the outset. Early approaches, such as SNIP \cite{lee2019snip} and GraSP \cite{wang2020picking}, leverage sensitivity metrics based on gradients to determine which weights can be safely pruned. Nevertheless, when applied globally, such methods suffer from layer collapse. In response, \citet{tanaka2020pruning} developed an iterative method of PaI, which mirrors that described in the previous paragraph but without re-training. This reduced layer-collapse occurrence, and improved the performance when PaI. However, recent results have suggested that the performance of such methods is not due to the selected nodes, but rather the per-layer PRs. As demonstrated by \citet{frankle2020pruning} and \citet{su2021prioritized}, the performance of models established using SNIP, GraSP, and SynFlow is robust to the weight shuffling within layers. Nevertheless, this phenomenon was not repeated at ultra-high sparsity. \citet{pham2024towards} argued that this was evidence that, when one aims to PaI, their objective should be to preserve the number of effective paths, as achieved in PHEW and NpB \cite{patil2021phew,pham2024towards}. These methods outperformed SNIP despite being data-independent. Nevertheless, they failed to attain results comparable to PaT, unlike ProsPr \cite{alizadeh2022prospect}.

\section{Mutual Information Preserving Pruning at a Glance}\label{sec:ps}

In this section, we introduce the required notation before a formal definition of MIPP.
The function describing a layer $l$ in NN can be written as follows: $\textup{f}_l(\boldsymbol{x}^n_{l}) = \boldsymbol{x}^m_{l+1} = \textup{a}(\boldsymbol{W}^{m \times n}_l\boldsymbol{x}^{n}_l + \boldsymbol{b}^m_l) .$ In the above, $\textup{a}$ is an activation function, $\boldsymbol{W}^{m \times n}_l$ is a weight matrix, $\boldsymbol{b}^m_l$ the bias, and $\boldsymbol{x}^{n}_l$ is the input to that layer \citep{lecun1998gradient,goodfellow2016deep}. 

Structured pruning is the process of discovering per-layer binary vector masks ($\boldsymbol{m}^{ n}_l$) that zero out weight matrix elements corresponding to a node or filter index. We will denote a pruned layer with a prime symbol ($'$) \citep{fahlman1990cascade}. The set of all masks associated with a network is given by \( \mathcal{M}_0 \), while the function associated with a pruned layer can be written as: $f'_l(\boldsymbol{x}^n_{l}) = \boldsymbol{x}'^m_{l+1} = \textup{a}(\boldsymbol{W}^{m \times n}_l\boldsymbol{x}^{n}_l\boldsymbol{m}^{ n}_l + \boldsymbol{b}^m_l)$. By randomly sampling from the space of possible inputs and applying the function described by the NN, we form not only the inputs as random variables (RVs), but also all subsequent activations. We define $X^i_l$ as the RV associated with the activations of node $i$ in layer $l$. Meanwhile, the set $\mathcal{X}_l = \{X^0_l, X^1_l\dots X^n_l\} $ contains a RV for all of the $N$ neurons in layer $l$. We use $\mathcal{X}_0$ to indicate the input. If a pruning mask is multiplied with the weights, the activations associated with pruned nodes are set to zero, which can otherwise be seen as information theoretically null. We denote the set associated with a pruned layer as $\mathcal{X}'_l$. If multiple pruning runs are performed with different datasets, multiple pruning masks will be created. In this case, both our pruning mask and our data distribution can be viewed as RVs,  $\mathcal{M}_0$ becoming $\mathcal{M}$, while $\mathcal{X}_0$ becomes $\mathcal{D}$. For a full table of notation please refer to Appendix \ref{app:notation}.

MIPP is founded on the idea that maximizing MI between realizations of the pruning mask and the data distribution, denoted as $I(\mathcal{D};\mathcal{M})$, ensures effective pruning with minimal performance loss. To achieve this, MIPP preserves MI between adjacent layers throughout a network. More specifically, we aim to isolate masks $\boldsymbol{m}^{ n}_l$, which combine with the weights to produce updated layers with some of the activations equal to zero. These null activations should not cause a reduction in the mutual information (MI) between the activations of adjacent layers.
More formally, this can be expressed as follows: 
$\mathcal{M}_0 = \{\boldsymbol{m}^{ n}_l  \forall l \in [1,L] : I(\mathcal{X}'_{l-1};\mathcal{X}_{l}) = I(\mathcal{X}_{l-1};\mathcal{X}_{l}) \}$.


\section{Theoretical Motivation} \label{sec:ps_mot}
We now motivate MIPP theoretically. As stated, we aim to design a method that preserves MI between activations such that $I(\mathcal{X}'_{l-1};\mathcal{X}_{l}) = I(\mathcal{X}_{l-1};\mathcal{X}_{l})$. In this section, we point out two advantages of doing this. Pruning in this manner not only ensures re-trainability, but it also leads to an optimal upper bound on the value of MI between the data-distribution and the masks $I(\mathcal{D};\mathcal{M})$ (as defined in \cite{kumar2024no}).

\textbf{Re-trainability.} We consider one-shot pruning with (re)-training: the objective is to reduce the number of nodes of the NN such that, after retraining, the pruned NN will achieve the same performance as the original.
We argue that one way to achieve this would be to select a subset of nodes from each layer in such a way that there exists a function, which, when applied to this subset, can still reconstruct the activations of the subsequent layer. We will then prove that the existence of this function preserves MI between the activations of these layers.

To illustrate this, we guide the reader through the following example. Consider the case in which we generate the expected outputs of our NN from the activations of the last layer. More formally, we write $\mathcal{X}_L = \textup{f}_{L-1}(\mathcal{X}_{L-1})$. We now wish to prune the activations preceding the outputs. This entails minimizing the number of nodes or the cardinality of the set $\mathcal{X}'_{L-1}$ in such a manner that there exists a function that can reliably re-form $\mathcal{X}_L$. Furthermore, this function should be discoverable by gradient ascent. More formally, we would like to derive $\mathcal{X}'_{L-1}$ such that $\mathcal{X}_L = \sup_{\textup{g} \in \mathcal{F}}\textup{g}(\mathcal{X}'_{L-1})$. While this formulation reveals little in the way of a potential pruning operation, using the following theorem, we relate it to the MI-based objective: $I(\mathcal{X}'_{l-1};\mathcal{X}_{l}) = I(\mathcal{X}_{l-1};\mathcal{X}_{l})$.

\noindent \textbf{Theorem 1:} \textit{There exists a function $\textup{g}$ such that the activations of the subsequent layer can be re-formed from the pruned layer iff MI between these two layers is not affected by pruning. More formally: $\mathcal{X}_L = \sup_{\textup{g} \in \mathcal{F}}\textup{g}(\mathcal{X}'_{L-1})\Leftrightarrow I(\mathcal{X}'_{L-1};\mathcal{X}_L) = I(\mathcal{X}_{L-1};\mathcal{X}_L)$.}

\noindent \textit{Proof.} See Appendix \ref{app:pot1}. \\ \noindent 
Consequently, in this work we aim to select a set of masks ($\mathcal{M}_0$) that increase sparsity while preserving MI between layers. This ensures that, for each pruned layer, there exists a function, discoverable by gradient descent, that effectively reconstructs the activations of the subsequent layer using those of the pruned layer. 
In other words, MIPP ensures re-trainability. We note that other methods demonstrate their ability to preserve retrainability through empirical results only \cite{wang2023trainability}.

\textbf{Maximizing $I(\mathcal{D};\mathcal{M}$).}
As discussed in the introduction, a sensible pruning objective is to maximize $I(\mathcal{D};\mathcal{M}$). \citet{westphal24} proved that TERC, the method we use to preserve MI between layers, does so via the derivation of a bijective function. This implies that the activations of the upstream pruned layer $\mathcal{X}'_{l-1}$ can be used to produce the downstream layer $\mathcal{X}_l$ and vice versa. In this section, we present theoretical results showing that the existence of such a bijective function allows the derivation of a maximum upper bound on the achievable MI between the masks and datasets $I(\mathcal{D};\mathcal{M})$.

\noindent \textbf{Theorem 2:} \textit{If a pruning method preserves MI between layers activations then the upper bound on $I(\mathcal{D};\mathcal{M})$ reaches its maximum. More formally: $I(\mathcal{X}'_{l-1};\mathcal{X}_{l}) = I(\mathcal{X}_{l-1};\mathcal{X}_{l}) \Leftrightarrow I(\mathcal{D};\mathcal{M}) \leq H(\mathcal{D})$.}

\noindent \textit{Proof.} See Appendix \ref{app:pot2}. \\ \noindent 
As a result, when using MIPP there is a greater upper-bound on the value of $I(\mathcal{D};\mathcal{M})$, which has been shown to be related to the models' accuracy and sample efficiency.
\section{MIPP}
\subsection{Preliminaries}

\subsubsection{Transfer Entropy Redundancy Criterion with MI Ordering} \label{sec:terc}
Before describing the method, we now provide a summary of TERC and its application to pruning, through the incorporation of an additional step for MI-based ordering.

\noindent \textbf{Node Pruning using TERC.} 
MIPP uses the transfer entropy redundancy criterion (TERC) \citep{westphal24} to dynamically prune nodes whose activations do not transfer entropy to the downstream layer. As discussed in Section \ref{sec:ps}, we aim to preserve MI between the layers in our network. The problem of MI preservation is one well-studied in the feature selection community \citep{battiti94,peng05,gao16}. 
We chose TERC, as not only does it preserve MI with the target via a bijective function, but its temporal complexity is also linear in time with respect to the number of features \citep{westphal24}, a key property when working in highly dimensional feature spaces. In our case, rather than selecting features to describe a target, we are selecting nodes that transfer entropy to the following layer. Within this context, TERC can be summarized as follows: to begin, all nodes in the layer are assumed to be useful (and added to the non-pruned set). We then sequentially evaluate whether the reduction in uncertainty of the subsequent layer's activations is greater when a specific node is included in the unpruned set rather than excluded. More formally, for a node $X^i_l$ to be added the set of pruned nodes, it must satisfy the following condition $I(\mathcal{X}_{l-1} ; \mathcal{X}_{l}) = I(\mathcal{X}_{l-1} \backslash X^i_l ; \mathcal{X}_{l})$. Otherwise, it is maintained in the network structure. This process is sequentially repeated for all nodes in the layer. As shown in \citet{westphal24}, this simple technique will preserve MI between layers. 
\newline
\noindent \textbf{MI Ordering.} Before applying TERC, we sort the nodes in the pruning layer in descending order of MI with the target (see Algorithm \ref{alg:algorithm} in Appendix \ref{app:alg}). This step is motivated by Theorem 3 in \citet{westphal24}. In particular, they prove that TERC alone selects unnecessary variables if there exists perfectly redundant variable subsets of different cardinalities. Ordering addresses this problem.


\subsubsection{Mutual Information Estimation}\label{sec:miest}
Unless restricting oneself to scenarios inapplicable to real-world data (e.g. discrete RVs), verifying the condition in Section \ref{sec:terc} is computationally intractable. Consequently, we must approximate the condition using MI estimates, for which many methods have been developed \citep{Moon1995,Paninski2007, belghazi2018,oord2018,Poole2019}. 

For the purposes of pruning, our MI estimates need to only be considered for comparison. Rather than using a method that is able to provide highly accurate estimates slowly \citep{franzese2024minde}, we require one that emphasizes speed and consistent results. For these reasons, we adopt the technique presented in \citet{covert2020understanding}, in which the authors demonstrate that MI between two random processes ($X$ and $Y$) can be approximated as the reduction in error estimation caused by using $X$ to predict $Y$. More formally: 
$I(X;Y) \approx \mathbb{E}[\textup{l}(\textup{f}_\emptyset(\emptyset), Y)] -  \mathbb{E}[\textup{l}(\textup{f}_X(X), Y)],$
where each $\textup{f}$ is some function approximated via loss $\textup{l}$. If the variables are discrete, and a cross entropy loss is used, then this value is exactly equal to the ground truth MI \citep{gadgil2023estimating}. Even if the variables are continuous and a mean squared error loss is used, the above value approaches MI under certain circumstances \citep{covert2020understanding}. 
\begin{figure*}[htbp]
    \vskip 0.2in
    \centering
    \includegraphics[width=\textwidth]{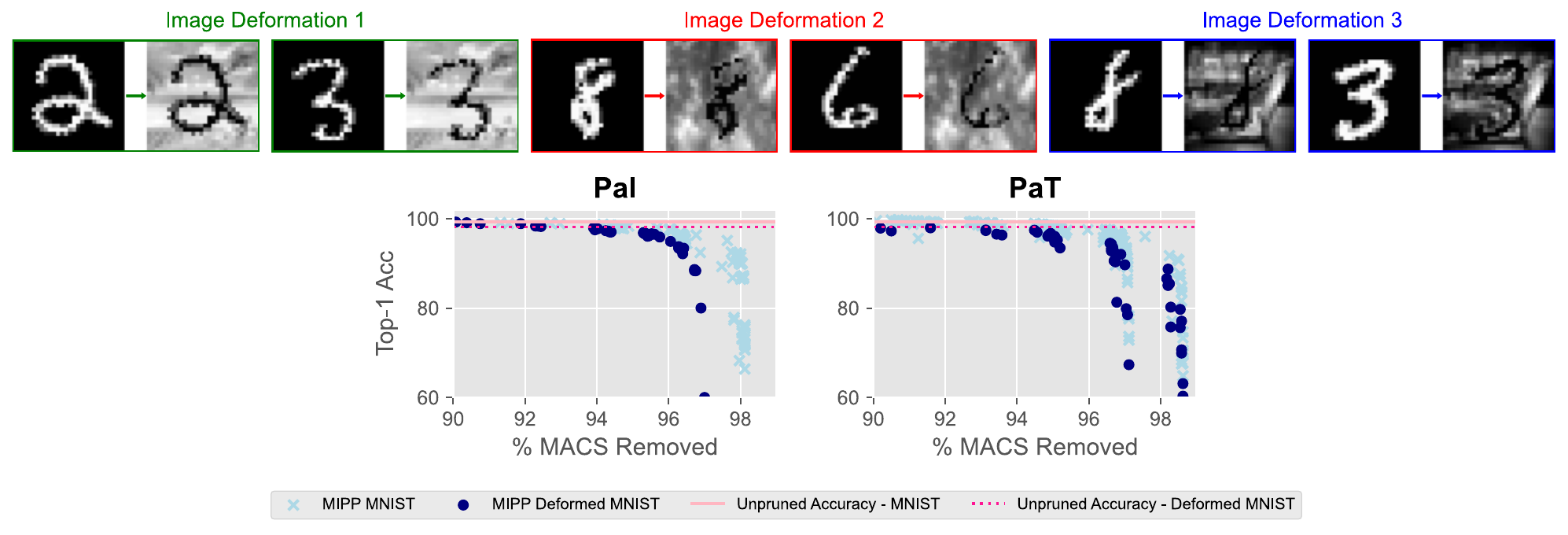}
    \caption{\textit{Top.} Deforming MNIST for increased image complexity. These transformations were applied randomly with equal probability and then kept consistent during training, pruning, and re-training. \textit{Bottom.} Changes in pruning ability of MIPP caused by image deformation.}
    \label{fig:deform_res}
    \vskip -0.2in
\end{figure*}

\begin{figure*}[!htbp]
    \vskip 0.2in
    \centering
    \includegraphics[width=\textwidth]{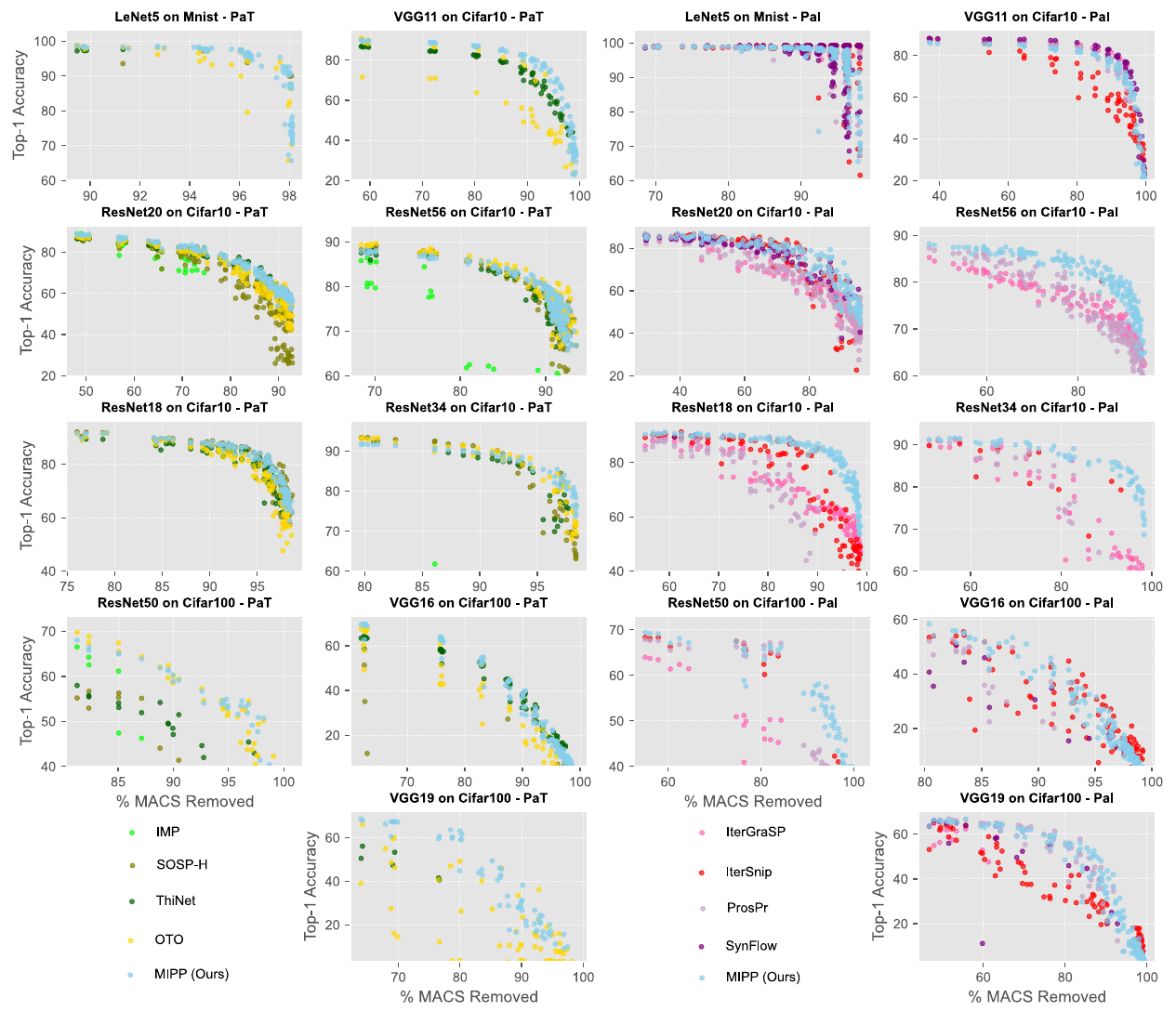}
    \caption{Comaprison of MIPP's ability to prune versus baselines both at initialization and after training. For clarity, we set an accuracy range to avoid viewing data points in which layer collapse has occurred.}
    \label{fig:main}
    \vskip -0.2in
\end{figure*}

\begin{figure*}[htbp]
    \vskip 0.2in
    \centering
    \includegraphics[width=\textwidth]{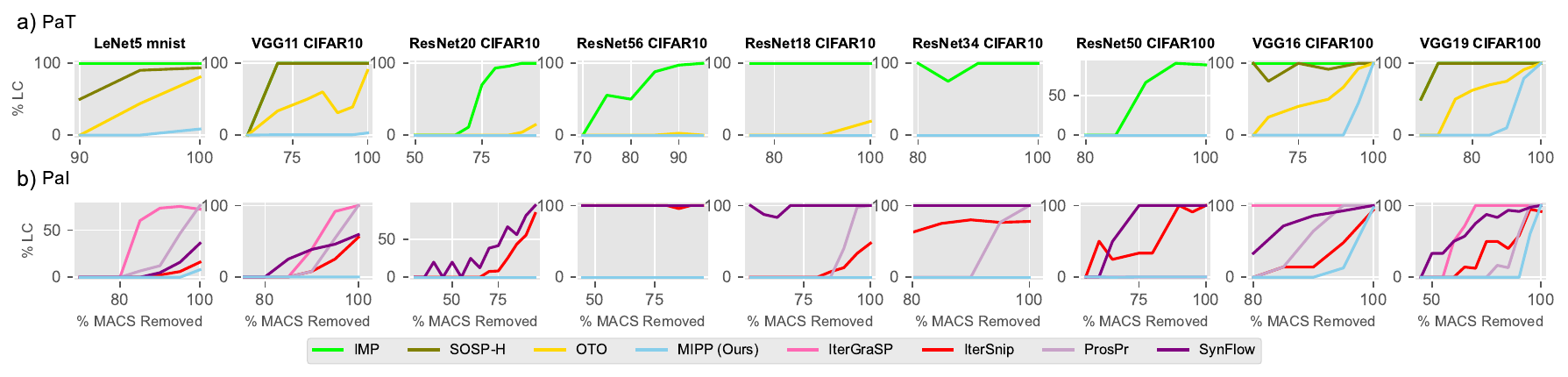}
    \caption{The percentage of runs that led to untrainable layer collapse. Specifically, we bin runs by the percentage of neurons removed, where one bin contains all the runs within a 5\% increment. We then calculate the percentage of these runs that lead to layer collapse.}
    \label{fig:lc}
    \vskip -0.2in
\end{figure*}

\begin{figure*}[htbp]
    \vskip 0.2in
    \centering
    \includegraphics[width=\textwidth]{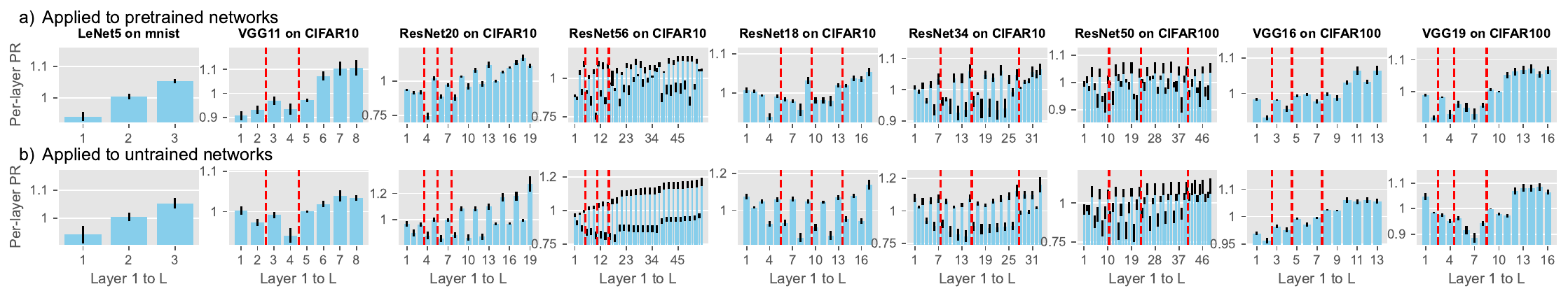}
    \caption{These experiments demonstrate the per-layer PR selected by MIPP. For the different layer-wise PRs we divide them by the average of all the layers in order to normalize. }
    \label{fig:nas}
    \vskip -0.2in
\end{figure*}

\subsection{Preserving the Mutual Information Between Adjacent Layers in Practice}
In this section, we discuss how to use TERC to preserve MI between a pair of adjacent layers.
As discussed, TERC with MI ordering dictates that, to remove a node, the following should be satisfied: $I(\mathcal{X}_{L-1} \backslash {X}^i_{L-1} ;\mathcal{X}_{L}) = I(\mathcal{X}_{L-1} ;\mathcal{X}_{L})$. In Section \ref{sec:miest}, we described the method we used to estimate MI. By combining these representations, we can update the condition we wish to approximate as follows:
\begin{equation}
\begin{split}
\label{eqn:covertterc}
I(\mathcal{X}_{l-1} ; \mathcal{X}_{l}) &= I(\mathcal{X}_{l-1} \backslash X^i_{l-1} ; \mathcal{X}_{l}) \\ & \text{(original condition as in TERC)}, \\
\mathbb{E}[\textup{l}(\textup{f}_{l-1}(\mathcal{X}_{l-1}), \mathcal{X}_{l})] &\geq \mathbb{E}[\textup{l}(\textup{h}_{l-1}(\mathcal{X}_{l-1} \setminus X^i_{l-1}), \mathcal{X}_{l})] \\ & \text{(estimated condition)}.
\end{split}
\end{equation} 
Equation \ref{eqn:covertterc} represents the simplification possible when $I(X;Y) \approx \mathbb{E}[\textup{l}(\textup{f}(\emptyset), Y)] -  \mathbb{E}[\textup{l}(\textup{f}(X), Y)]$ is substituted into $I(\mathcal{X}_{l-1} ; \mathcal{X}_{l}) = I(\mathcal{X}_{l-1}\backslash X^i_{l-1} ; \mathcal{X}_{l})$. Our condition characterizes the case where node $X^i_{l-1}$ transfers no entropy to the following layer. The monotonicity of MI enforces that, in this case, we have the equality seen in line one of Equation \ref{eqn:covertterc}. When approximating this condition, as shown in the third line of Equation \ref{eqn:covertterc}, we can no longer guarantee monotonicity. Therefore, we relax the equality to the inequality as indicated. Overall, our condition becomes a simple comparison of two losses quantifying two functions' ability to reconstruct the downstream layer. The definition of MI as presented in \citet{covert2020understanding} is applicable for any $\textup{f}_{l-1}$ or $\textup{h}_{l-1}$ discovered using function approximation. However, we need not fit a new function as we already posses $\textup{f}_{l-1}$ exactly in the form of layer $l$ in our network. Meanwhile, $\textup{h}_{l-1}$ is that same function but re-trained to predict the downstream layer's activations with node $X^i_{l}$ masked in its input. If $\mathbb{E}[\textup{l}(\textup{h}_{l-1}(\mathcal{X}_{l-1} \setminus X^i_{l-1}), \mathcal{X}_{l})]$ is equal to or drops below $\mathbb{E}[\textup{l}(\textup{f}_{l-1}(\mathcal{X}_{l-1}), \mathcal{X}_{l})]$ our condition is satisfied and we can remove the node $X^i_{l-1}$.


Given the above condition, we now describe TERC with MI ordering: initially, we order the nodes in descending order of the loss achieved when using just this variable as input to predict the downstream layer $\textup{f}_l(X_l^i)$. We now sequentially traverse the nodes in this order, similarly to \cite{gadgil2023estimating}, masking them and re-training our layer (to find $\textup{h}_l(\mathcal{X}_l \backslash X_l^i)$) to determine whether the loss function drops back below its original value $\mathbb{E}[\textup{l}(\textup{f}_{l-1}(\mathcal{X}_{l-1}), \mathcal{X}_{l})] \geq \mathbb{E}[\textup{l}(\textup{h}_{l-1}(\mathcal{X}_{l} \setminus X^i_{l-1}), \mathcal{X}_{l})]$. If it fails to recover, this implies that, without the activations of this node, we are unable to reconstruct the activations of the downstream layer. In this case, the variable is considered informative and should be retained in the network and in the set $\mathcal{X}'_{l-1}$. Otherwise, the node is removed. Once a node has been evaluated, the layer can be updated with the new trained function ($\textup{h}_{l-1}$).

We have explained how MIPP is a structured pruning method that retains nodes whose activations transfer entropy to the next layer. The number of nodes maintained in the network is therefore dynamically dependent on those already selected, making us unable to set a pruning ratio in the traditional sense \cite{hassibi1992second,nonnemacher2022sosp}. However, we wish to study MIPP at different degrees of sparsity. Consequently, we now briefly explain how we affect the pruning ratio discovered using our method. From the condition above, it is clear that for a node to be removed from the network, the loss must fall below the level achieved using all the activations, $\mathbb{E}[\textup{l}(\textup{f}_{l-1}(\mathcal{X}_{l-1}), \mathcal{X}_{l})]$.To adjust the pruning ratio, we update this threshold by allowing it to take values that are regularly spaced within the range $ \left[\mathbb{E}[\textup{l}(\textup{f}_{l-1}(\mathcal{X}_{l-1}), \mathcal{X}_{l})], \mathbb{E}[\textup{l}(\textup{f}_{l-1}(\emptyset), \mathcal{X}_{l})]\right]$. If we are close to $\mathbb{E}[\textup{l}(\textup{f}_{l-1}(\emptyset), \mathcal{X}_{l})]$, the condition for removing nodes is easily satisfied, and the sparsity ratio is high.


\subsection{Preserving the Mutual Information of the Entire Model from Outputs to Inputs } Until now, we have focused on the use of TERC with MI ordering to preserve MI between the activations of adjacent layers. This process is repeated for each pair of layers.
To prune the entire model, by preserving MI between pairs of layers, one could start from the input layer and move to the output layer or vice versa. In this section, like \citet{luo2017thinet}, we argue for the latter option, providing both theoretical and practical arguments.


\textbf{Theoretical argument.} Because each layer in an NN is an injective function of its predecessor, these pairs share perfect MI. In this case $I(\mathcal{X}_{l-1};\mathcal{X}_{l}) = H(\mathcal{X}_{l})$ \citep{cover1999elements}. Therefore, the networks layers can only reduce in entropy from inputs to outputs \citep{tishby2015information,shwartz2017opening}. Suppose we take the first approach, pruning from inputs to outputs. Our goal is to prune the first layer (\(\mathcal{X}_{1}\)), such that the result can be used to reconstruct the activations of the second layer (\(\mathcal{X}_{2}\)).
Since the second layer has not yet been pruned, it may retain redundant information, which is then maintained in the activations of the first layer during pruning. In contrast, if we take the second approach, we begin by pruning the activations in layer $\mathcal{X}_{L-1}$. The information in $\mathcal{X}'_{L-1}$ (its pruned version) has been preserved due its ability to reconstruct exclusively the outputs. Upon moving onto the next pair, we prune layer $\mathcal{X}_{L-2}$ based on the entropy in the layer $\mathcal{X}'_{L-1}$. Notably though, this has already been reduced by the first pairwise pruning step. According to this recursive logic, even when pruning the first layer, we are only retaining the entropy required to reproduce the outputs.

\textbf{Practical argument.} We now present a more practical reason for pruning from outputs to inputs rather than vice versa. Under this scheme we aim to evaluate the condition $I(\mathcal{X}'_{l-1};\mathcal{X}'_{l}) = I(\mathcal{X}_{l-1};\mathcal{X}'_{l})$, rather than $I(\mathcal{X}'_{l-1};\mathcal{X}_{l}) = I(\mathcal{X}_{l-1};\mathcal{X}_{l})$ which would be appropriate if forward pruning was conducted. In the former case, we apply our function to predict a layer whose dimensionality has already been reduced. This increases efficiency by mitigating the effects of the curse of dimensionality \citep{bellman1959}. Algorithm \ref{alg:mipp_algorithm} describes the steps of MIPP formally.  Notably, the utility of MIPP can extend beyond just pruning. 
By verifying which pixels transfer entropy to the activations of the pruned first layer, MIPP also demonstrates the ability to select features. The corresponding experiments are presented in Appendix \ref{app:fs_expts}.

\section{Evaluation}
In this section, we discuss the evaluation of MIPP, starting from the experimental settings and the datasets. We selected MNIST, CIFAR-10, and CIFAR-100 for their benchmark status, and diversity, enabling comprehensive evaluation of MIPP while ensuring comparability \cite{lecun1989optimal,krizhevsky2009learning}.

\subsection{Models, datasets and baselines} 
We begin by applying our method to the simple LeNet5 architecture detecting variations of the MNIST dataset \citep{lecun1998gradient}. We then assess its ability to prune ResNet20, ResNet56, ResNet18, ResNet34, and VGG11 on the CIFAR10 dataset \citep{he2016deep,simonyan2014very}. Before then investigating ResNet50,  VGG16, and VGG19 models networks trained on CIFAR100 \citep{krizhevsky2009learning}. When using MIPP to PaI, we compare to SynFlow \citep{tanaka2020pruning}, IterGraSP \citep{wang2020picking}, IterSNIP (both 100 iterations) and ProsPr. Meanwhile, when using MIPP to PaT we compare to IMP \cite{frankle2018lottery}, OTO \cite{you2020drawing},  ThiNet \citep{luo2017thinet} and SOSP-H \citep{nonnemacher2022sosp}. GraSP, SynFlow and SNIP are unstructured; in order to make them structured, we apply L1-normalization to all the weights associated with a node. MIPP selects nodes based on whether their activations transfer entropy to those of the subsequent layer. This approach inherently establishes a unique PR for each run, which we adopt as the global PR for our baseline methods. ThiNet cannot determine layer-wise PR; therefore, we apply a uniform PR across all layers. 

\subsection{LeNet5 on varieties of MNIST} In this section, we begin by discussing how increasing the dataset complexity of MNIST affects MIPP's results, we then examine more generally how MIPP performs compared to baselines as applied to a LeNet5 on MNIST.

Empirical evidence indicates that the utility of PaI may be limited to simple datasets \cite{frankle2018lottery,frankle2020pruning}. In these experiments, we intend to provide an information-theoretic argument as to why this might be, with supporting empirical evidence. MIPP, and other effective PaI schemes, preserve and compress the information encoded in network activations. In untrained networks, these activations reflect the \textit{entirety} of the information present in the input data. If these inputs are characterized by information relevant to the classification task, MIPP (and PaI more generally) remains applicable. For instance, in the MNIST dataset, the informative pixels assist the classification task, while the remaining pixels, on the outskirts of the image, are constantly black and contain no information. In such cases, our method selectively preserves the neurons whose activations correspond to informative pixels. On the other hand, the converse is also true; our method is inapplicable to models whose input data contains information not relevant for the classification task. Consequently, if the input data is complex, the ability to prune at initialization is reduced. To demonstrate this effect, in Figure \ref{fig:deform_res} we present experiments that investigate the effects of deforming MNIST. In alignment with our hypothesis, we observe a reduction in our ability to prune an untrained network but not a trained network with increased data complexity. 

Figure \ref{fig:main} demonstrates that MIPP performs at least as well as the baselines, regardless of whether PaI or PaT. Additionally, Figure \ref{fig:lc} shows that our method exhibits greater resistance to layer collapse.

\subsection{Other Models on CIFAR10/100}
In Figure \ref{fig:main}, it is clear that MIPP performs at least as well as baselines, whether PaI or PaT, on most models when either CIFAR10 or CIFAR100 are acting as input. When pruning ResNet18 and ResNet34 at initialization MIPP, for high sparsities, outperforms baselines by over 15\%. We observe similarly impressive results when pruning a VGG19 trained on CIFAR100. These results demonstrate that certain global pruning objectives can be used to PaI or PaT. In Figure \ref{fig:lc} it is clear that MIPP is consistently the most resistant to layer collapse for all model dataset combinations. Meanwhile, in Figure \ref{fig:nas} we observe MIPP forms highly regularized layer-wise sparsity ratios, particularly when applied at initialization.  In particular, we observe block-based PRs, as well as intra-block PR patterns for ResNets. In Appendix \ref{app:resnets} we provide a pictorial explanation of the ResNet structure, from which it is possible to observe the periodicities of the intra-block structure in a ResNet34 and 50.

\section{Conclusion}
In this paper, we have introduced MIPP, an activation-based pruning method that can be applied both before and after training. The core principle of MIPP is to remove neurons or filters from layers if they do not transfer entropy to the subsequent layer. Consequently, MIPP preserves MI between the activations of adjacent layers and, therefore, between the data and masks. We have presented a comprehensive performance evaluation of MIPP considering a variety of datasets and models. Our experimental evaluation has demonstrated the effectiveness of MIPP in pruning trained and untrained models of increasing complexity.

\section*{Impact Statement}
This work investigates network pruning as a method to reduce the computational cost and energy consumption of deep learning models, thereby contributing to the environmental sustainability of AI technologies. We believe that MIPP could have a positive impact if deployed, and we do not foresee any specific issues with the proposed technology. Code will be made available on publication.


\bibliography{example_paper}

\begin{thebibliography}{50}
\providecommand{\natexlab}[1]{#1}
\providecommand{\url}[1]{\texttt{#1}}
\expandafter\ifx\csname urlstyle\endcsname\relax
  \providecommand{\doi}[1]{doi: #1}\else
  \providecommand{\doi}{doi: \begingroup \urlstyle{rm}\Url}\fi

\bibitem[Alizadeh et~al.(2022)Alizadeh, Ajanthan, Yuan, and Torr]{alizadeh2022prospect}
Alizadeh, M., Ajanthan, T., Yuan, X., and Torr, P.~H.
\newblock Prospect pruning: Finding trainable weights at initialization using meta-gradients.
\newblock In \emph{ICLR'22}, 2022.

\bibitem[Battiti(1994)]{battiti94}
Battiti, R.
\newblock Using mutual information for selecting features in supervised neural net learning.
\newblock \emph{IEEE Transactions on Neural Networks}, 5\penalty0 (4):\penalty0 537--550, 1994.

\bibitem[Belghazi et~al.(2018)Belghazi, Rajeswar, Baratin, Hjelm, and Courville]{belghazi2018}
Belghazi, I., Rajeswar, S., Baratin, A., Hjelm, R.~D., and Courville, A.
\newblock Mine: Mutual information neural estimation.
\newblock In \emph{ICML'18}, 2018.

\bibitem[Bellman \& Kalaba(1959)Bellman and Kalaba]{bellman1959}
Bellman, R. and Kalaba, R.
\newblock A mathematical theory of adaptive control processes.
\newblock \emph{Proceedings of the National Academy of Sciences}, 45\penalty0 (8):\penalty0 1288--1290, 1959.

\bibitem[Blalock et~al.(2020)Blalock, Gonzalez~Ortiz, Frankle, and Guttag]{blalock2020state}
Blalock, D., Gonzalez~Ortiz, J.~J., Frankle, J., and Guttag, J.
\newblock What is the state of neural network pruning?
\newblock In \emph{MLSys'20}, 2020.

\bibitem[Cover(1999)]{cover1999elements}
Cover, T.~M.
\newblock \emph{Elements of Information Theory}.
\newblock John Wiley \& Sons, 1999.

\bibitem[Covert et~al.(2020)Covert, Lundberg, and Lee]{covert2020understanding}
Covert, I., Lundberg, S.~M., and Lee, S.-I.
\newblock Understanding global feature contributions with additive importance measures.
\newblock In \emph{NeurIPS'20}, 2020.

\bibitem[DeVries \& Taylor(2017)DeVries and Taylor]{devries2017cutout}
DeVries, T. and Taylor, G.~W.
\newblock Improved regularization of convolutional neural networks with cutout.
\newblock In \emph{arXiv preprint arXiv:1708.04552}, 2017.

\bibitem[Fahlman \& Lebiere(1990)Fahlman and Lebiere]{fahlman1990cascade}
Fahlman, S.~E. and Lebiere, C.
\newblock The cascade-correlation learning architecture.
\newblock \emph{NeurIPS'90}, 1990.

\bibitem[Frankle \& Carbin(2019)Frankle and Carbin]{frankle2018lottery}
Frankle, J. and Carbin, M.
\newblock The lottery ticket hypothesis: Finding sparse, trainable neural networks.
\newblock In \emph{ICLR'19}, 2019.

\bibitem[Frankle et~al.(2021)Frankle, Dziugaite, Roy, and Carbin]{frankle2020pruning}
Frankle, J., Dziugaite, G.~K., Roy, D.~M., and Carbin, M.
\newblock Pruning neural networks at initialization: Why are we missing the mark?
\newblock In \emph{ICLR'21}, 2021.

\bibitem[Franzese et~al.(2024)Franzese, Bounoua, and Michiardi]{franzese2024minde}
Franzese, G., Bounoua, M., and Michiardi, P.
\newblock {MINDE}: Mutual information neural diffusion estimation.
\newblock In \emph{ICLR'24}, 2024.

\bibitem[Gadgil et~al.(2024)Gadgil, Covert, and Lee]{gadgil2023estimating}
Gadgil, S., Covert, I., and Lee, S.-I.
\newblock Estimating conditional mutual information for dynamic feature selection.
\newblock In \emph{ICLR'24}, 2024.

\bibitem[Gao et~al.(2016)Gao, Ver~Steeg, and Galstyan]{gao16}
Gao, S., Ver~Steeg, G., and Galstyan, A.
\newblock Variational information maximization for feature selection.
\newblock In \emph{NeurIPS'16}, 2016.

\bibitem[Goodfellow et~al.(2016)Goodfellow, Bengio, Courville, and Bengio]{goodfellow2016deep}
Goodfellow, I., Bengio, Y., Courville, A., and Bengio, Y.
\newblock \emph{Deep Learning}.
\newblock MIT Press, 2016.

\bibitem[Han et~al.(2015)Han, Pool, Tran, and Dally]{han2015learning}
Han, S., Pool, J., Tran, J., and Dally, W.~J.
\newblock Learning both weights and connections for efficient neural networks.
\newblock In \emph{NeurIPS'15}, 2015.

\bibitem[Hassibi \& Stork(1992)Hassibi and Stork]{hassibi1992second}
Hassibi, B. and Stork, D.
\newblock Second order derivatives for network pruning: Optimal brain surgeon.
\newblock In \emph{NeurIPS'92}, 1992.

\bibitem[He et~al.(2016)He, Zhang, Ren, and Sun]{he2016deep}
He, K., Zhang, X., Ren, S., and Sun, J.
\newblock Deep residual learning for image recognition.
\newblock In \emph{CCVPR'16}, 2016.

\bibitem[Krizhevsky(2009)]{krizhevsky2009learning}
Krizhevsky, A.
\newblock \emph{Learning multiple layers of features from tiny images}.
\newblock PhD thesis, University of Toronto, 2009.

\bibitem[Kumar et~al.(2024)Kumar, Luo, and Sellke]{kumar2024no}
Kumar, T., Luo, K., and Sellke, M.
\newblock No free prune: Information-theoretic barriers to pruning at initialization.
\newblock In \emph{ICML'24}, 2024.

\bibitem[LeCun et~al.(1989)LeCun, Denker, and Solla]{lecun1989optimal}
LeCun, Y., Denker, J., and Solla, S.
\newblock Optimal brain damage.
\newblock In \emph{NeurIPS'89}, 1989.

\bibitem[LeCun et~al.(1998)LeCun, Bottou, Bengio, and Haffner]{lecun1998gradient}
LeCun, Y., Bottou, L., Bengio, Y., and Haffner, P.
\newblock Gradient-based learning applied to document recognition.
\newblock \emph{Proceedings of the IEEE}, 86\penalty0 (11):\penalty0 2278--2324, 1998.

\bibitem[Lee et~al.(2019)Lee, Ajanthan, and Torr]{lee2019snip}
Lee, N., Ajanthan, T., and Torr, P.~H.
\newblock {SNIP: Single-shot network pruning based on connection sensitivity}.
\newblock In \emph{ICLR'19}, 2019.

\bibitem[Li et~al.(2017)Li, Kadav, Durdanovic, Samet, and Graf]{li2017pruning}
Li, H., Kadav, A., Durdanovic, I., Samet, H., and Graf, H.~P.
\newblock Pruning filters.
\newblock In \emph{ICLR'17}, 2017.

\bibitem[Liu et~al.(2017)Liu, Li, Shen, Huang, Yan, and Zhang]{liu2017learning}
Liu, Z., Li, J., Shen, Z., Huang, G., Yan, S., and Zhang, C.
\newblock Learning efficient convolutional networks through network slimming.
\newblock In \emph{ICCV'17}, 2017.

\bibitem[Luo et~al.(2017)Luo, Wu, and Lin]{luo2017thinet}
Luo, J.-H., Wu, J., and Lin, W.
\newblock Thinet: A filter level pruning method for deep neural network compression.
\newblock In \emph{ICCV'17}, 2017.

\bibitem[Molchanov et~al.(2017)Molchanov, Tyree, Karras, Aila, and Kautz]{molchanov2017pruning}
Molchanov, P., Tyree, S., Karras, T., Aila, T., and Kautz, J.
\newblock Pruning convolutional neural networks for resource efficient inference.
\newblock In \emph{ICLR'17}, 2017.

\bibitem[Molchanov et~al.(2019)Molchanov, Mallya, Tyree, Frosio, and Kautz]{molchanov2019importance}
Molchanov, P., Mallya, A., Tyree, S., Frosio, I., and Kautz, J.
\newblock Importance estimation for neural network pruning.
\newblock In \emph{CCVPR'19}, 2019.

\bibitem[Moon et~al.(1995)Moon, Rajagopalan, and Lall]{Moon1995}
Moon, Y.-I., Rajagopalan, B., and Lall, U.
\newblock Estimation of mutual information using kernel density estimators.
\newblock \emph{Physical Review E}, 52\penalty0 (3):\penalty0 2318--2321, 1995.

\bibitem[Nonnenmacher et~al.(2022)Nonnenmacher, Pfeil, Steinwart, and Reeb]{nonnemacher2022sosp}
Nonnenmacher, M., Pfeil, T., Steinwart, I., and Reeb, D.
\newblock {SOSP: Efficiently Capturing Global Correlations by Second-Order Structured Pruning}.
\newblock In \emph{ICLR'22}, 2022.

\bibitem[Paninski(2003)]{Paninski2007}
Paninski, L.
\newblock Estimation of entropy and mutual information.
\newblock \emph{Neural Computation}, 15\penalty0 (6):\penalty0 1191–1253, 2003.

\bibitem[Patil \& Dovrolis(2021)Patil and Dovrolis]{patil2021phew}
Patil, S.~M. and Dovrolis, C.
\newblock Phew: Constructing sparse networks that learn fast and generalize well without training data.
\newblock In \emph{ICML'21}, 2021.

\bibitem[Peng et~al.(2005)Peng, Long, and Ding]{peng05}
Peng, H., Long, F., and Ding, C.
\newblock Feature selection based on mutual information criteria of max-dependency, max-relevance, and min-redundancy.
\newblock \emph{IEEE Transactions on Pattern Analysis and Machine Intelligence}, 27\penalty0 (8):\penalty0 1226--1238, 2005.

\bibitem[Peng et~al.(2019)Peng, Wu, Chen, and Huang]{peng2019collaborative}
Peng, H., Wu, J., Chen, S., and Huang, J.
\newblock Collaborative channel pruning for deep networks.
\newblock In \emph{ICML'19}, 2019.

\bibitem[Pham et~al.(2024)Pham, Liu, Xiang, Le, Wen, Tran-Thanh, et~al.]{pham2024towards}
Pham, H., Liu, S., Xiang, L., Le, D., Wen, H., Tran-Thanh, L., et~al.
\newblock Towards data-agnostic pruning at initialization: what makes a good sparse mask?
\newblock In \emph{NeurIPS'24}, 2024.

\bibitem[Poole et~al.(2019)Poole, Ozair, Oord, Alemi, and Tucker]{Poole2019}
Poole, B., Ozair, S., Oord, A., Alemi, A., and Tucker, G.
\newblock On variational bounds of mutual information.
\newblock In \emph{ICML'19}, 2019.

\bibitem[Shwartz-Ziv \& Tishby(2017)Shwartz-Ziv and Tishby]{shwartz2017opening}
Shwartz-Ziv, R. and Tishby, N.
\newblock Opening the black box of deep neural networks via information.
\newblock \emph{arXiv preprint arXiv:1703.00810}, 2017.

\bibitem[Simonyan \& Zisserman(2015)Simonyan and Zisserman]{simonyan2014very}
Simonyan, K. and Zisserman, A.
\newblock Very deep convolutional networks for large-scale image recognition.
\newblock In \emph{ICLR'15}, 2015.

\bibitem[Su et~al.(2021)Su, Huang, Li, You, Wang, Qian, Zhang, and Xu]{su2021prioritized}
Su, X., Huang, T., Li, Y., You, S., Wang, F., Qian, C., Zhang, C., and Xu, C.
\newblock {Prioritized Architecture Sampling with Monte-Carlo Tree Search}.
\newblock In \emph{CCVPR'21}, 2021.

\bibitem[Sung et~al.(2024)Sung, Yoon, and Bansal]{sung2024ecoflap}
Sung, Y.-L., Yoon, J., and Bansal, M.
\newblock {ECoFLaP: Efficient Coarse-to-Fine Layer-Wise Pruning for Vision-Language Models}.
\newblock In \emph{ICLR'24}, 2024.

\bibitem[Tanaka et~al.(2020)Tanaka, Kunin, Yamins, and Ganguli]{tanaka2020pruning}
Tanaka, H., Kunin, D., Yamins, D.~L., and Ganguli, S.
\newblock Pruning neural networks without any data by iteratively conserving synaptic flow.
\newblock In \emph{NeurIPS'20}, 2020.

\bibitem[Tishby \& Zaslavsky(2015)Tishby and Zaslavsky]{tishby2015information}
Tishby, N. and Zaslavsky, N.
\newblock Deep learning and the information bottleneck principle.
\newblock In \emph{ITW'15}, 2015.

\bibitem[van~den Oord et~al.(2019)van~den Oord, Li, and Vinyals]{oord2018}
van~den Oord, A., Li, Y., and Vinyals, O.
\newblock Representation learning with contrastive predictive coding.
\newblock In \emph{arXiv:1807.03748}, 2019.

\bibitem[Wang et~al.(2019)Wang, Grosse, Fidler, and Zhang]{wang2019eigendamage}
Wang, C., Grosse, R., Fidler, S., and Zhang, G.
\newblock {EigenDamage: Structured pruning in the Kronecker-factored eigenbasis}.
\newblock In \emph{ICML'19}, 2019.

\bibitem[Wang et~al.(2022)Wang, Zhang, and Grosse]{wang2020picking}
Wang, C., Zhang, G., and Grosse, R.
\newblock Picking winning tickets before training by preserving gradient flow.
\newblock In \emph{ICLR'22}, 2022.

\bibitem[Wang \& Fu(2023)Wang and Fu]{wang2023trainability}
Wang, H. and Fu, Y.
\newblock Trainability preserving neural pruning.
\newblock In \emph{ICLR'23}, 2023.

\bibitem[Westphal et~al.(2024)Westphal, Hailes, and Musolesi]{westphal24}
Westphal, C., Hailes, S., and Musolesi, M.
\newblock Information-theoretic state variable selection for reinforcement learning.
\newblock \emph{arXiv preprint arXiv:2401.11512}, 2024.

\bibitem[You et~al.(2020)You, Li, Xu, Fu, Wang, Chen, Baraniuk, Wang, and Lin]{you2020drawing}
You, H., Li, C., Xu, P., Fu, Y., Wang, Y., Chen, X., Baraniuk, R.~G., Wang, Z., and Lin, Y.
\newblock Drawing early-bird tickets: Toward more efficient training of deep networks.
\newblock In \emph{ICLR'20}, 2020.

\bibitem[Zhang et~al.(2018)Zhang, Cisse, Dauphin, and Lopez-Paz]{zhang2018mixup}
Zhang, H., Cisse, M., Dauphin, Y.~N., and Lopez-Paz, D.
\newblock mixup: Beyond empirical risk minimization.
\newblock In \emph{ICLR'18}, 2018.

\bibitem[Zhao et~al.(2019)Zhao, Ni, Zhang, Zhao, Zhang, and Tian]{zhao2019variational}
Zhao, C., Ni, B., Zhang, J., Zhao, Q., Zhang, W., and Tian, Q.
\newblock Variational convolutional neural network pruning.
\newblock In \emph{CVPR'19}, 2019.

\end{thebibliography}
\bibliographystyle{icml2025}

\newpage
\appendix
\onecolumn
\newpage

\section{Notation}\label{app:notation}
\begin{table}[th] 
\centering
\caption{Summary of Notational Conventions}
\label{tab:notation}
\begin{tabular}{@{}ll@{}} 
    \toprule
    \textbf{Type}                                & \textbf{Notation}                           \\ \midrule
    Loss function                                    & $\textup{l}$ \\
    The function describing layer $l$                                    & $\textup{f}_l$ \\
    The function describing layer $l$ once masked and re-trained                                   & $\textup{h}_l$ \\
    A vector of activations associated with layer l                                 & $\boldsymbol{x}^{n}_l$ \\
    RV describing the activations of node i in layer l                                     & $X^i_l$ \\

    Set of RVs describing all activations in layer l (one realization being $\boldsymbol{x}^{n}_l$)                                & $\mathcal{X}_l$ \\
    Image vector                                 & $\boldsymbol{x}^{n}_0$ \\
    
    Set of pixel RVs describing input data (one realization would be an image $\boldsymbol{x}^{n}_0$)                                 & $\mathcal{X}_0$ \\
    RV describing different datasets (one realization would be an dataset such as MNIST or $\mathcal{X}_0$)                                 & $\mathcal{D}$ \\
    The mask vector used for layer l                               & $\boldsymbol{m}^{n}_l$ \\
    The set of masks established for all layers in a model                               & $\mathcal{M}_0$ \\
    The RV describing the different masks established for different data (one realization: $\mathcal{M}_0$)                               & $\mathcal{M}$ \\
    Set of RVs describing pruned activations in layer l  & $\mathcal{X}'_l$ \\
    RV describing different pruned activations in layer l occurring due to different datasets (realization:$\mathcal{X}'_l$)  & $\mathcal{D}'_l$ \\
    Weight matrix                                     & $\boldsymbol{W}^{m \times n}$ \\
    Bias vector                                     & $\boldsymbol{b}^{m }$ \\
    Activation function                                     & $\textup{a}$ \\       \bottomrule
\end{tabular}
\end{table}
\section{Algorithms}\label{app:alg}
In this section, we present not only the overall MIPP algorithm but also TERC with MI ordering algorithm, which maintains MI between adjacent layers in a network.
\begin{algorithm}
    \caption{MIPP.}
    \label{alg:mipp_algorithm}
    \textbf{Input}: Activations of all layers: $\mathcal{X}_{l}$. \textbf{Output}: $\mathcal{M}_0$ (a desirable set of node masks).
    \begin{algorithmic}[1]
    \STATE Initialize empty set of masks: $\mathcal{M}_0 = \emptyset$.
    \FOR{$l \in [1, L]$}
        \STATE ${\mathcal{X}}'_{l-1}= \text{Algorithm \ref{alg:algorithm}}(\mathcal{X}_{l-1}, \mathcal{X}_{l})$
        \FOR{$i \in [0, n]$}
        \STATE $\boldsymbol{m}^n_{l-1}(i) = 
\begin{cases} 
1 & \text{if } X^i_{l-1} \in \mathcal{X}'_{l-1}, \\ 
0 & \text{otherwise}.

\end{cases}
$
\ENDFOR
\STATE $\mathcal{M}_0=\mathcal{M}_0 \cup \boldsymbol{m}^n_{l-1}$
    \ENDFOR

    \STATE \textbf{return} $\mathcal{M}_0$
    \end{algorithmic}
\end{algorithm}
\begin{algorithm}[th]
    \caption{TERC with MI ordering.}
    \label{alg:algorithm}
    \textbf{Input}: Activations of layers $L$ and $L-1$: $\mathcal{X}_{L}$ and $\mathcal{X}_{L-1}$. \textbf{Output}: ${\mathcal{X}}'_{L-1}$ (a desirable subset of nodes).
    \begin{algorithmic}[1]
    \STATE Initialize ${\mathcal{X}}'_{L-1} = \text{sort}_{\text{desc}}\left( \mathcal{X}_{L-1}, I({X}^i_{L-1};\mathcal{X}_{L}) \right)$
    \FOR{${X}^i_{L-1} \in \mathcal{X}_{L-1}$}
        \IF{$I(\mathcal{X}'_{L-1} \backslash {X}^i_{L-1} ;\mathcal{X}_{L}) = I(\mathcal{X}_{L-1} ;\mathcal{X}_{L})$} 
                \STATE ${\mathcal{X}'_{L-1}} = \mathcal{X}'_{L-1} \backslash {X}^i_{L-1}$
        \ENDIF
    \ENDFOR
    \STATE \textbf{return} ${\mathcal{X}}'_{L-1}$
    \end{algorithmic}
\end{algorithm}

\section{Proof of Theorem 1}\label{app:pot1}

In this section we prove Theorem 1. To begin, we remind the reader that we aim to preserve MI between layers such that:
\begin{equation}
\begin{split}
\label{eqn:pot1}
I(\mathcal{X}'_{L-1};\mathcal{X}_L) & = I(\mathcal{X}_{L-1};\mathcal{X}_L), 
\end{split}
\end{equation}
which, given the relationship $I(X;Y)=\sup_f \left( \mathbb{E}[f(X \mid Y] - \log \mathbb{E}[e^{f(X)}] \right)$, becomes:
\begin{equation}
\begin{split}
\label{eqn:pot12}
 \sup_g \left( \mathbb{E}[g(\mathcal{X}_{L-1}) \mid \mathcal{X}_L] - \log \mathbb{E}[e^{g(\mathcal{X}_{L-1})}] \right) =& \sup_f \left( \mathbb{E}[f(\mathcal{X}'_{L-1}) \mid \mathcal{X}_L] - \log \mathbb{E}[e^{f(\mathcal{X}'_{L-1})}] \right).
\end{split}
\end{equation}
However, we know that there exists a function $g$ such that $g(\mathcal{X}_{L-1}) = \mathcal{X}_L$. Therefore, we can rewrite the above such that:
\begin{equation}
\begin{split}
\label{eqn:pot13}
 \left( \mathbb{E}[\mathcal{X}_L \mid \mathcal{X}_L] - \log \mathbb{E}[e^{\mathcal{X}_L}] \right) =& \sup_f \left( \mathbb{E}[f(\mathcal{X}'_{L-1}) \mid \mathcal{X}_L] - \log \mathbb{E}[e^{f(\mathcal{X}'_{L-1})}] \right), \\
 \mathcal{X}_L - \log \mathbb{E}[e^{\mathcal{X}_L}] =& \sup_f \left( \mathbb{E}[f(\mathcal{X}'_{L-1}) \mid \mathcal{X}_L] - \log \mathbb{E}[e^{f(\mathcal{X}'_{L-1})}] \right). \\
\end{split}
\end{equation}
The only circumstances under which Equation \ref{eqn:pot12} holds is if $f(\mathcal{X}'_{L-1}) = \mathcal{X}_L$, thereby proving Theorem 1.

\section{Proof of Theorem 2}\label{app:pot2}
In this section, we prove Theorem 2. To begin, we present axioms that will be used throughout the proof. 

\begin{itemize}
    \item Firstly, we apply TERC to bijectively preserve MI between activations in layers such that $I(\mathcal{X}'_{l-1};\mathcal{X}_{l}) = I(\mathcal{X}_{l-1};\mathcal{X}_{l})$. Given that $\mathcal{X}_{l} = \textup{f}(\mathcal{X}_{l-1})$, this implies that from the pruned upstream layer we should be able to perfectly reconstruct the original layer $I(\mathcal{X}_l;\mathcal{X}'_{l-1})= H(\mathcal{X}_{l-1})$ \cite{westphal24}.
    \item With probability one can we recover the mask if we have access to the masked activations $p(m|\mathcal{X}'_{l-1}) = 1.$
    \item With probability one can we recover the masked activations if we have access to the full activations $p(\mathcal{X}'_{l-1}| \mathcal{X}_{l-1}) = 1.$
    \item We assume that all the information in the data is included in the first layer of the activations $I(\mathcal{X}_0;\mathcal{X}_1)=H(\mathcal{X}_0)$. Finally, for this proof, we also assume a network with one set of activations to prune. 
\end{itemize}
To begin, we remind the reader that $\mathcal{D}$ is a distribution from which we sample input data. Therefore, an instance of $\mathcal{D}$ can be written as the input data to our NN, written $\mathcal{X}_0$. Meanwhile, $\mathcal{M}$ is a RV whose realizations are the sets of masks derived using a pruning method, denoted $\mathcal{M}_0$. We use these observations to complete the proof.
\begin{align}
\label{eqn:pot2}
I(\mathcal{D}; \mathcal{M}) &= \sum_{\mathcal{X}_0 \in \mathcal{D}, \mathcal{M}_0 \in \mathcal{M}} p(\mathcal{X}_0, \mathcal{M}_0) \log \left( \frac{p(\mathcal{X}_0, \mathcal{M}_0)}{p(\mathcal{X}_0) \, p(\mathcal{M}_0)} \right) \\
& \text{(substituting in $p(\mathcal{X}_1|\mathcal{X}_0, \mathcal{M}_0) = 1$ we obtain)} \\
&= \sum_{\mathcal{X}_0 \in \mathcal{D}, \mathcal{M}_0 \in \mathcal{M}} p(\mathcal{X}_0, \mathcal{M}_0) \log \left( \frac{p(\mathcal{X}_0, \mathcal{M}_0, \mathcal{X}_1)}{p(\mathcal{X}_0) \, p(\mathcal{M}_0)} \right) \\
& \text{(substituting in $p(\mathcal{X}'_1|\mathcal{X}_0, \mathcal{M}_0, \mathcal{X}_1) = 1$ we obtain)} \\
&= \sum_{\mathcal{X}_0 \in \mathcal{D}, \mathcal{M}_0 \in \mathcal{M}} p(\mathcal{X}_0, \mathcal{M}_0) \log \left( \frac{p(\mathcal{X}_0, \mathcal{M}_0, \mathcal{X}_1, \mathcal{X}'_1)}{p(\mathcal{X}_0) \, p(\mathcal{M}_0)} \right) \\
&= \sum_{\mathcal{X}_0 \in \mathcal{D}, \mathcal{M}_0 \in \mathcal{M}} p(\mathcal{X}_0, \mathcal{M}_0) \log \left( \frac{p(\mathcal{M}_0 | \mathcal{X}_0, \mathcal{X}_1, \mathcal{X}'_1)p(\mathcal{X}_0, \mathcal{X}_1, \mathcal{X}'_1)}{p(\mathcal{X}_0) \, p(\mathcal{M}_0)} \right) \\
& \text{(because $p(\mathcal{M}_0|\mathcal{X}_1, \mathcal{X}'_1) = 1$ we obtain)} \\
&= \sum_{\mathcal{X}_0 \in \mathcal{D}, \mathcal{M}_0 \in \mathcal{M}} p(\mathcal{X}_0, \mathcal{M}_0) \log \left( \frac{p(\mathcal{X}_0, \mathcal{X}_1, \mathcal{X}'_1)}{p(\mathcal{X}_0) \, p(\mathcal{M}_0)} \right) \\
&= \sum_{\mathcal{X}_0 \in \mathcal{D}, \mathcal{M}_0 \in \mathcal{M}} p(\mathcal{X}_0, \mathcal{M}_0) \log \left( \frac{p(\mathcal{X}'_1 | \mathcal{X}_0, \mathcal{X}_1)p(\mathcal{X}_0, \mathcal{X}_1)}{p(\mathcal{X}_0) \, p(\mathcal{M}_0)} \right) \\
& \text{(because $p(\mathcal{X}'_1 | \mathcal{X}_0, \mathcal{X}_1) = 1$ we obtain)} \\
&= \sum_{\mathcal{X}_0 \in \mathcal{D}, \mathcal{M}_0 \in \mathcal{M}} p(\mathcal{X}_0, \mathcal{M}_0) \log \left( \frac{p(\mathcal{X}_0, \mathcal{X}_1)}{p(\mathcal{X}_0) \, p(\mathcal{M}_0)} \right) \\
&= \sum_{\mathcal{X}_0 \in \mathcal{D}, \mathcal{M}_0 \in \mathcal{M}} p(\mathcal{X}_0, \mathcal{M}_0) \log \left( \frac{p(\mathcal{X}_1|\mathcal{X}_0)p(\mathcal{X}_0)}{p(\mathcal{X}_0) \, p(\mathcal{M}_0)} \right) \\
& \text{(because $p(\mathcal{X}_1 | \mathcal{X}_0) = 1$ we obtain)} \\
&= \sum_{\mathcal{X}_0 \in \mathcal{D}, \mathcal{M}_0 \in \mathcal{M}} p(\mathcal{X}_0, \mathcal{M}_0) \log \left( \frac{1}{ p(\mathcal{M}_0)} \right) \\
&= H(\mathcal{M}) \\
\end{align}

As previously pointed out, an instance of our masking variable $\mathcal{M}$ is a single set of masks $\mathcal{M}_0$. It is clearly possible to derive this mask from $\mathcal{X}'_1$; therefore, we obtain $\mathcal{M}_0 = \textup{f}(\mathcal{X}'_1)$ and $H(\mathcal{M}) \leq  H(\mathcal{D}'_1)$ (where $\mathcal{D}'_1$ is the RV from which $\mathcal{X}'_1$ is sampled). 

We can then write:
\begin{align}
\label{eqn:pot22}
H(\mathcal{M}) & \leq  H(\mathcal{D}'_1)\\
&\leq -\sum_{\mathcal{X}'_1 \in \mathcal{D}'_1} p(\mathcal{X}'_1) \log p(\mathcal{X}'_1) \\
& \text{(because of TERC's bijective MI preservation we can sub in $p(\mathcal{X}_1 | \mathcal{X}'_1) = 1$)} \\
&\leq -\sum_{\mathcal{X}'_1 \in \mathcal{D}'_1} p(\mathcal{X}'_1, \mathcal{X}_1) \log p(\mathcal{X}'_1, \mathcal{X}_1) \\
&\leq -\sum_{\mathcal{X}'_1 \in \mathcal{D}'_1} p(\mathcal{X}'_1 | \mathcal{X}_1) p(\mathcal{X}_1) \log p(\mathcal{X}'_1 | \mathcal{X}_1) p(\mathcal{X}_1) \\
& \text{(by repeating the process above but inserting $p(\mathcal{X}_0 | \mathcal{X}_1) = 1$)} \\
& \leq  H(\mathcal{D}). \\
\end{align}
We have proven that, if pruning using a method that bijectively preserves MI between pruned and unpruned activations, the upper bound on $I(\mathcal{D};\mathcal{M})$ can be expressed as $I(\mathcal{D};\mathcal{M}) \geq H(\mathcal{D})$. 

\section{Further Experimental Settings}
\subsection{$I(\mathcal{D};\mathcal{M})$ vs Top-1 Acc}\label{app:linei}
In this section, we explain the experimental settings for the results achieved in Figure \ref{fig:obj} a). In these simple introductory experiments, we aimed to provide empirical evidence validating the results of \citet{kumar2024no} in different settings. To achieve this, we used synthetic data to generate a mask for which we could calculate $I(\mathcal{D};\mathcal{M})$, and then applied this mask to all layers of the network. We would then train this network (controlling for the initialized weight matrices) and present the final accuracy seen in the figure.

\textbf{Data.} In this case, to simplify the process of deriving our masks $\mathcal{M}_0$, our synthetic data was an $N$-dimensional vector of Bernoulli distributions. Of these $N$ Bernoulli distributions, half were described by $\sim \text{Bernoulli}(0.5)$, while the other half were described by $\sim \text{Bernoulli}(0.999999)$. We denote this vector as $\textbf{d}^N$. To generate masks with high MI with the data, they should accurately reflect the patterns in the input data. For instance, if we are generating a set of masks with perfect MI with the data, and the informative Bernoulli distributions occupy the first 25 positions, then our mask will have its first 25 positions set to prune. Meanwhile, if the informative Bernoulli distributions appeared once in every other array element, we may repeat this with the positions we mask. What matters for perfect MI is that the masks are a perfect function of the distribution vector, represented as $\mathcal{M}_0 = f(\textbf{d}^N)$. In our case, this function is simply a 1-to-1 mapping. To reduce MI, we simply add randomness. For instance, if we aimed to reduce the MI by half, we would have only have half our masking vector be dependent on the data, while the rest is random. Finally, for these experiments, the target (i.e., $y$) is a simple sum of all the inputs.

The output data we were trying to predict was a sum of the inputs.

\textbf{Models.} For our simple feed-forward neural network (NN), we used 50 inputs and 3 hidden layers, each containing 50 nodes. In contrast, our convolutional neural network (ConvNet) had one convolutional layer with 1 input channel, 8 output channels, and a kernel size of 3. We then flattened the output and fed it into a linear network with 384 units.

\subsection{What PaI methods maximize $I(\mathcal{D};\mathcal{M})$?}\label{app:whatpai}
In this section, we explain the experimental settings for the results reported in Figure \ref{fig:obj} b). The goal of these experiments is to show that for a simple synthetic network characterized by redundancies, MIPP establishes masks that have a greater MI with the data. We employed a simple MLP composed of 10 hidden nodes, with input data that was also a vector of dimension 10, where, similarly to in Appendix \ref{app:linei}, this vector was made of samples from Bernoulli distributions, half of which were informative ($\sim \text{Bernoulli}(0.5)$) while the other half ($\sim \text{Bernoulli}(0.999999)$) were not. The network takes this input vector of zeroes and ones and converts them from a binary value to a decimal one. In this case, a perfect network will have a weight matrix that has powers of two along the diagonal. We then use multiple PaI methods to establish pruning masks. To calculate $I(\mathcal{D};\mathcal{M})$, we use the methods described in \cite{covert2020understanding} with a small network of two hidden layers with 50 nodes and BCE loss. 

We observe that only MIPP can define pruning masks that are data-dependent. This is due to the following reasons: firstly, the network performs perfectly and there is no loss, so the derivative-based metrics are zero; and, secondly, the activations give perfectly redundant information, masking them would have the exact same effect on the loss (assuming you use a cross entropy loss and not MSE). Under such conditions, static ranking methods produce results that resemble randomness.
 
\subsection{Data Augmentation Techniques}
For the CIFAR-10 dataset, we applied standard data augmentation techniques, which included random cropping with padding and random horizontal flipping. These augmentations are commonly used to enhance model generalization by introducing variations in the training data. In the case of CIFAR-100, we employed additional augmentation methods beyond the standard techniques. Specifically, we used mixup \citep{zhang2018mixup}, which creates virtual training examples by combining pairs of images and their labels, and cutout \citep{devries2017cutout}, which randomly masks out square regions of an image to simulate occlusion and encourage the network to focus on more distributed features. These advanced techniques were included to further enhance performance due to the increased complexity of the CIFAR-100 dataset.

\subsection{Hyperparameters}
\subsubsection{Vision Training and Re-training}
Please refer to Table \ref{tab:training-params}.
\begin{table}[th]
    \centering
    \caption{Comparison of training parameters across datasets.}
    \label{tab:training-params}
    \begin{tabularx}{\textwidth}{@{}Xlll@{}}
        \toprule
        \textbf{Dataset} & \textbf{MNIST} & \textbf{CIFAR10} & \textbf{CIFAR100} \\
        \midrule
        \textbf{Solver} & SGD (0.9, 1e-4) & SGD (0.9, 5e-4) & SGD (0.9, 5e-4) \\
        \textbf{Batch Size} & 100 & 128 & 256 \\
        \textbf{LR} & 1e-2, [30,60], \#epochs:90 & 1e-1, [100,150], \#epochs:200 & 1e-1, [100,150], \#epochs:200 \\
        \textbf{LR~(re-train)} & 1e-2, [30], \#epochs:60 & 1e-2, [60,90], \#epochs:120 & 1e-2, [60,90], \#epochs:120 \\
        \bottomrule
    \end{tabularx}
\end{table}

\subsubsection{Pruning}
As explained in the main paper, our method involves masking features and re-training a layer to check if the loss decreases below the original level. For the re-training steps once the mask has been applied we use 100 epochs with 20 batches of activations with early stopping. At the start of our algorithm, we also rank the features based of their MI. For this calculation, we again use the same layer for 5 epochs with the same 20 activations. 
We use our method to prune all linear and convolutional layers. We prune the batch-normalization nodes associated with nodes in linear/convolutional layers, while skip connections in ResNets remain unaffected. We note that the batch-normalization layer can natively be considered as part of the function $\textup{f}_{l-1}$ and therefore retrained to form $\textup{h}_{l-1}$. Consequently, MIPP considers batch-normalization layers to greater effect than other pruning methods. 
For OTO and IMP, we used 10 iterations with 20 retraining epochs. The learning rates are reported in Table \ref{tab:training-params}.

\newpage
\section{Further Experiments}
\begin{figure*}[b!]
    \centering
    \includegraphics[width=\columnwidth]{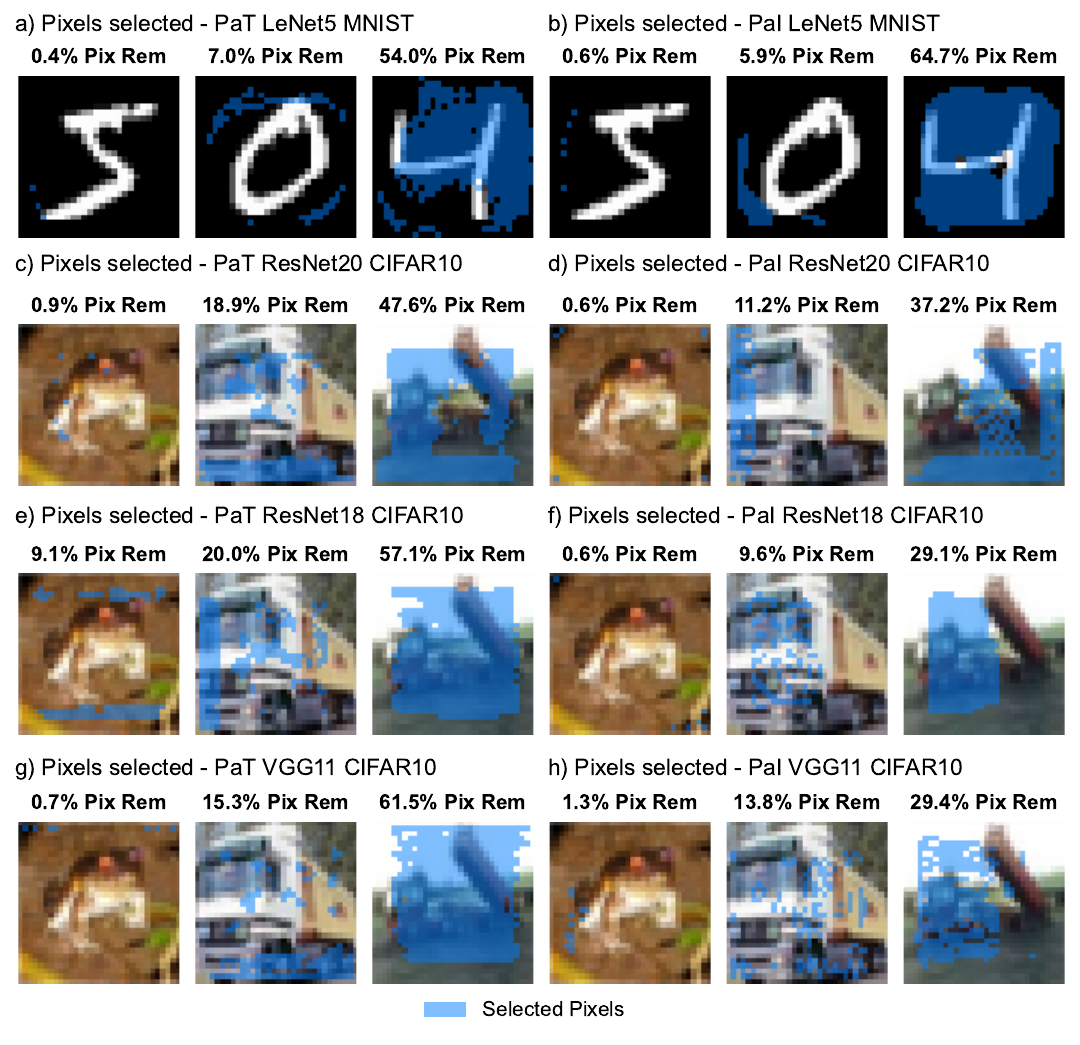}
    \caption{Visual representation of the features selected using MIPP at different sparsities on different models and datasets (blue implies selected).}
    \label{fig:selected_pix}
\end{figure*}
\begin{figure*}[t]
    \centering
    \includegraphics[width=\columnwidth]{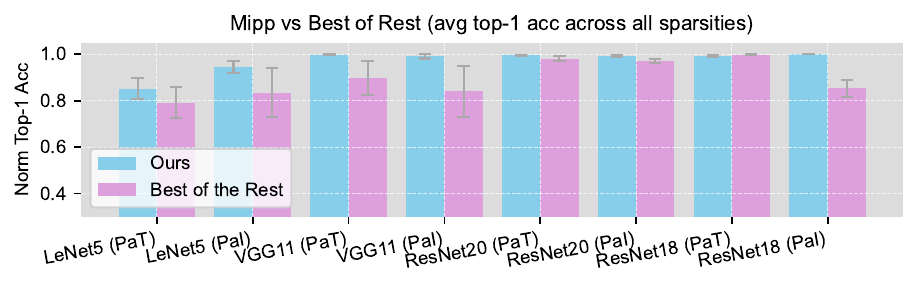}
    \caption{Normalized accuracy of MIPP vs best of the rest when pixel selection occurs.}
    \label{fig:pix_bars}
\end{figure*}

\subsection{Feature Selection Experiments }\label{app:fs_expts}
In this section, we investigate MIPP's ability to select features.
In particular, we examine the pixels it identifies from the MNIST and CIFAR10 datasets. MIPP selects features in the exact same manner it selects nodes, i.e., by verifying whether entropy is transferred from the pixels to the nodes of the first layer. In order to do this, we extract the first layer and rank the pixels based on their impact on reconstructing the activations of the next layer, from least to most significant. We then sequentially evaluate the pixels in this order, testing whether retraining the layer with each pixel masked results in a loss that is lower than or equal to the loss obtained using all the pixels.

In Figure \ref{fig:selected_pix}, we observe a clear preference for selecting central pixels over those located at the periphery, which are generally less informative. In panels (a) and (b) of Figure \ref{fig:selected_pix}, we present results that align with previous research on MNIST pixel importance \cite{covert2020understanding}. These studies have demonstrated that the most informative pixels are typically concentrated in the center of the images, with a slight bias toward the right side. Additionally, an intuitive selection pattern is evident in the CIFAR-10 dataset. When pixel selection is performed after training, the selected pixels exhibit a high degree of uniformity. In contrast, when selection occurs at initialization, regular patterns still emerge, although the spacing between selected pixels is less consistent. 

In Figure \ref{fig:pix_bars}, we present the average accuracy achieved when we prune models using MIPP and our baselines. Unlike the experiment reported in the main body of the paper, we have also used MIPP to select pixels. In both figures, we observe that MIPP outperforms the baselines. This is because, unlike any of the baselines, the features are selected in a manner that is dependent on the pruned model. MIPP can compress both features and the underlying model simultaneously such that the results are compatible,  preventing ML practitioners from having to use different methods for feature and model compression. Often, combining compressed input and compressed models can lead to performance degradation.

\newpage

\subsection{Layer Wise Pruning Ratios Established Using Other Methods}\label{app:layer_collapse}
In the main paper, we present the per-layer PRs obtained by MIPP. In Figure \ref{fig:nas}, we present these results for the other methods taken into consideration.

\begin{figure}[H]
    \centering
    \vskip 0.5in
    \includegraphics[width=\textwidth, height=0.85\textheight]{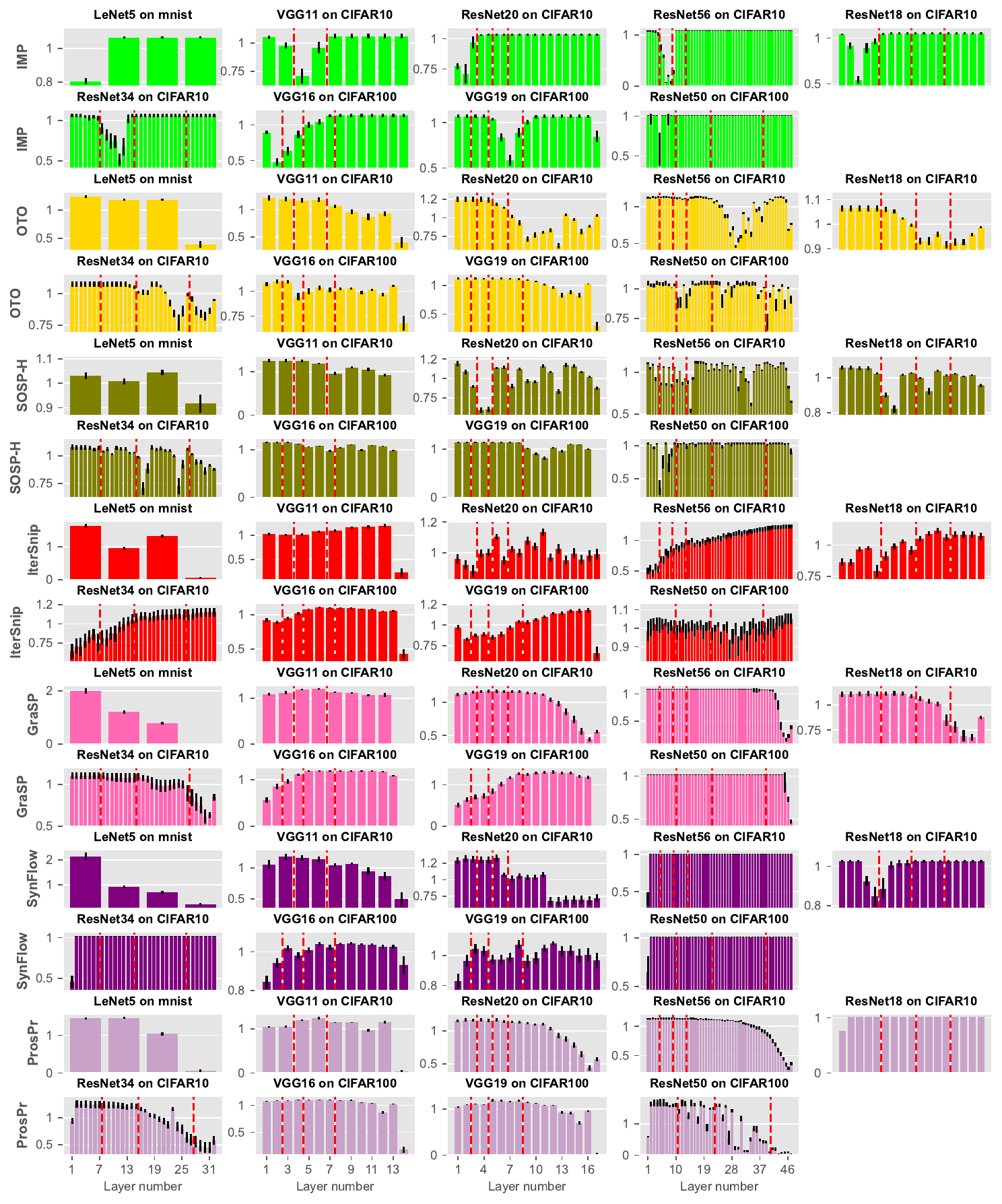}
    \caption{Layer-wise pruning ratios. Normalized by division of the average PR achieved for that run.}
    \label{fig:other_pl_pr}
    \vskip -0.5in
\end{figure}\newpage
\section{ResNet Structure}\label{app:resnets}
\begin{figure}[h!]
\vskip 0.2in
    \centering
    \includegraphics[width=0.5\columnwidth]{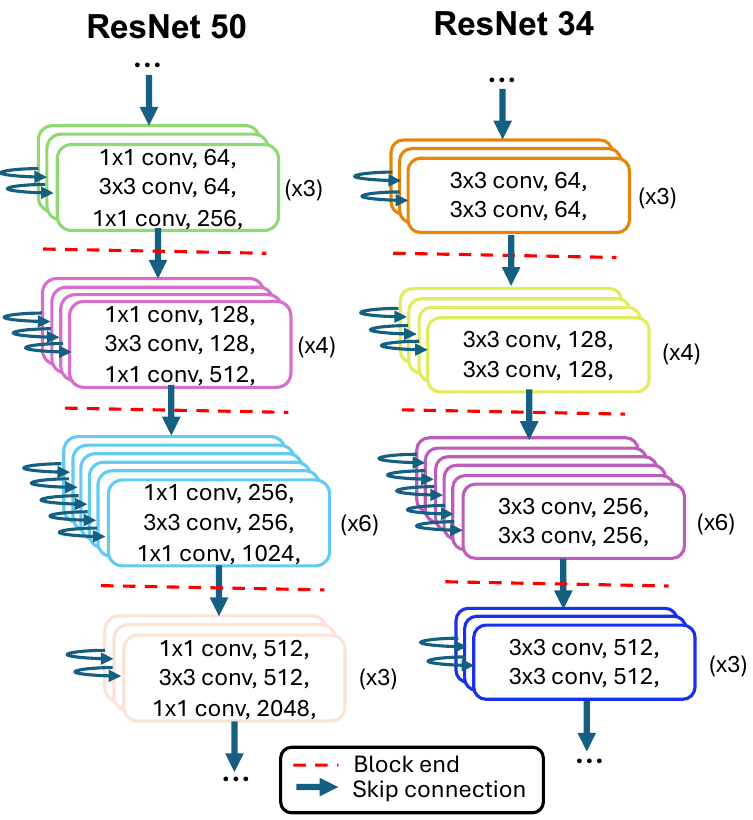}
    \caption{ResNet34 and ResNet50 structures: the architectural elements of the network explain the periodicity of the per-layer PRs derived using our method.}
    \label{fig:resnets}
\vskip -0.2in
\end{figure}

In this section, we present Figure \ref{fig:resnets}, which illustrates the structure of some of the ResNets investigated explaining the per-layer pruning ratios discovered in Figure \ref{fig:other_pl_pr}.


\end{document}